\documentclass[runningheads]{llncs}
\usepackage[T1]{fontenc}
\usepackage{subfigure}
\usepackage{mathtools}
\usepackage{graphicx}
\usepackage{booktabs}
\usepackage{multirow}
\usepackage{multicol}
\usepackage{amsmath}
\usepackage{amssymb}
\usepackage{bbding}
\usepackage{colortbl}
\usepackage{pifont}
\usepackage{CJKutf8}
\usepackage{bbm}
\usepackage{bm}
\usepackage{dsfont}
\usepackage{xcolor}
\definecolor{Gray}{gray}{0.93}
\def\@mb@citenamelist{cite,citep,citet,citealp,citealt,citepalias,citetalias}

\newcommand{\modelname}{\textsc{RoNID}}

\begin{document}
\begin{CJK*}{UTF8}{gbsn}
\title{\modelname{}: New Intent Discovery with Generated-Reliable Labels and Cluster-friendly Representations}
%
%
\author{
  Shun Zhang\textsuperscript{\rm 1,2\thanks{Equal contribution.}}, 
  Chaoran Yan\textsuperscript{\rm 1 $^\star$}, 
  Jian Yang\textsuperscript{\rm 1\thanks{Corresponding author.}},\\
  Changyu Ren\textsuperscript{\rm 1},
  Jiaqi Bai\textsuperscript{\rm 1,2},
  Tongliang Li\textsuperscript{\rm 3},
  Zhoujun Li\textsuperscript{\rm 1,2}\\
  \textsuperscript{\rm 1}State Key Lab of Software Development Environment, Beihang University \\
  \textsuperscript{\rm 2}School of Cyber Science and Technology, Beihang University \\
  \textsuperscript{\rm 3}Beijing Information Science and Technology University \\
  \{shunzhang,ycr2345,jiaya,cyren,bjq,lizj\}@buaa.edu.cn\\
\{tonyliangli\}@bistu.edu.cn
}

%

\institute{}
\maketitle              
\begin{abstract}
New Intent Discovery (NID) strives to identify known and reasonably deduce novel intent groups in the open-world scenario.  
But current methods face issues with inaccurate pseudo-labels and poor representation learning, creating a negative feedback loop that degrades overall model performance, including accuracy and the adjusted rand index.
To address the aforementioned challenges, we propose a \textbf{Ro}bust \textbf{N}ew \textbf{I}ntent \textbf{D}iscovery (\modelname{}) framework optimized by an EM-style method, which focuses on constructing reliable pseudo-labels and obtaining cluster-friendly discriminative representations.
\modelname{} comprises two main modules: \textit{reliable pseudo-label generation module} and \textit{cluster-friendly representation learning module}. 
Specifically, the pseudo-label generation module assigns reliable synthetic labels by solving an optimal transport problem in the \texttt{E}-step, which effectively provides high-quality supervised signals for the input of the cluster-friendly representation learning module.
To learn cluster-friendly representation with strong intra-cluster compactness and large inter-cluster separation, the representation learning module combines intra-cluster and inter-cluster contrastive learning in the \texttt{M}-step to feed more discriminative features into the generation module.
\modelname{} can be performed iteratively to ultimately yield a robust model with reliable pseudo-labels and cluster-friendly representations.
Experimental results on multiple benchmarks demonstrate our method brings substantial improvements over previous state-of-the-art methods by a large margin of +1$\sim$+4 points.

\keywords{New Intent Discovery  \and Reliable Pseudo-labels Generation \and Cluster-friendly Representation Learning.}
\end{abstract}
\section{Introduction}
Understanding and identification of user intent accurately are important to downstream task-oriented dialogue systems, which wield direct influence over both system performance and user perception~\cite{Dialog2,Dialog3}.
Traditional dialogue systems are based on the closed-world assumption \cite{chen2019bert}, resulting in their limited ability to only identify pre-defined and constrained intent classes.
Considering the limitation of the closed-world assumptions, new user intents inevitably emerge in the real-world scenario, which emphasizes the necessity of dynamically uncovering novel intents.
New intent discovery (NID) methods~\cite{lin2020discovering,zhang2021discovering,zhang-2022-new-intent-discovery,DPN,zhou2023latent} are proposed to infer novel user intents from narrow pre-defined labels and large-scale raw utterances.

\begin{figure}[t]   
\centering
\includegraphics[width=0.85\columnwidth]{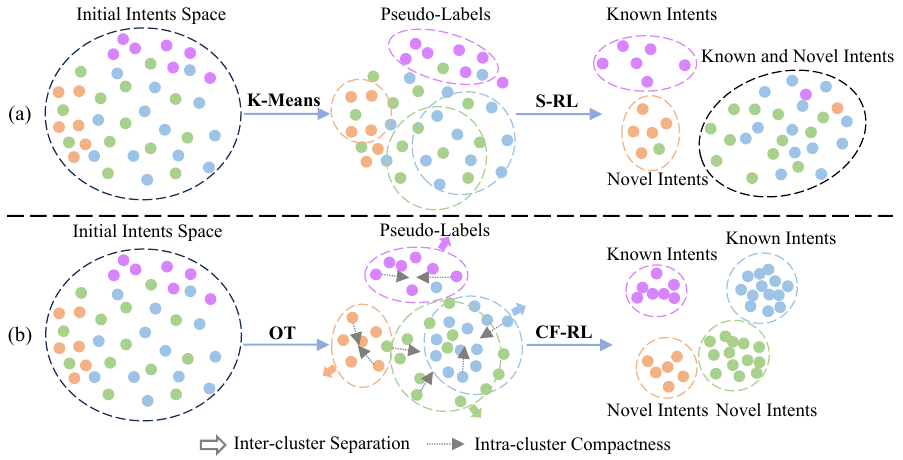}
\caption{
Illustration of the defects of baselines and the advantages of our method. Compared to the baselines (a) with the clustering degeneracy, our method (b) successfully separates known and novel intents based on reliable pseudo-labels and cluster-friendly representations. \texttt{S-RL} denotes suboptimal representation learning and \texttt{CF-RL} denotes cluster-friendly representation learning.
}
\label{fig:fig_1}
\end{figure}

Along the research line of unsupervised setting with clustering algorithms~\cite{padmasundari2018intent,min2020dialogue}, all samples are considered unlabeled during the clustering process to discover novel categories. The unsupervised methods often yield unsatisfactory performance due to the lack of guidance of prior knowledge.
Another line of efforts~\cite{DPN,zhou2023latent} resorts to the semi-supervised setting, which aims to discover intents by fully utilizing the evidence from limited labeled data for the effective cluster. Semi-supervised NID typically employs the k-means algorithm for pseudo-label assignment and learns discriminative intent features. Despite the development of semi-supervised NID, previous methods still encounter two challenges: (1) \textbf{unreliable pseudo-labels assignments}, attributed to the training noise and sensitivity of the k-means algorithm to outliers. (2) \textbf{suboptimal representation learning}, ignoring the explicit modeling of intra-class compactness and inter-class separability and failing to capture cluster-friendly representations. Both challenges form a feedback loop of self-reinforcing errors during pseudo-labeling causing severe clustering degeneracy. Therefore, two directions need further exploration. (\textbf{CH1}) \textit{How to design a reliable pseudo-label assignments method to construct high-quality guidance for representation learning?} and (\textbf{CH2}) \textit{How to obtain cluster-friendly discriminative intent representations for NID?}

To mitigate the above weakness, we propose a novel robust NID framework, \modelname{}, to boost the NID performance with a reliable pseudo-label generation module and a cluster-friendly representation learning module.
%
To tackle \textbf{CH1}, we address the problem of intent pseudo-label assignments through the paradigm of optimal transport, which is the first attempt in the NID domain.
We formulate the process of assigning pseudo-labels to unlabeled data as an optimal transport problem. By constraining the distribution of the generated pseudo-labels to match the estimated class distribution in \texttt{E}-step as closely as possible, the pseudo-labels gradually approximate the ground truth.
We encourage a uniform and dynamically updated estimated class distribution to prevent cluster degeneration while utilizing the Sinkhorn-Knopp algorithm~\cite{peyre2019computational,cuturi2013sinkhorn} to optimize the objective.
%
To address \textbf{CH2}, in the \texttt{M}-step, we incorporate both intra-cluster and inter-cluster contrastive learning objects into the cluster-friendly representation learning module.
The intra-cluster contrastive learning uses pseudo-labels as the supervision to compel instances to move closer to their corresponding prototype, achieving greater compactness within clusters.
While inter-cluster contrastive learning maximizes the distance between clusters from the prototype-to-prototype perspective, it aims to attain a larger separation between clusters.
The two modules cooperate with each other and the iterative training process forms a virtuous circle, where the pseudo-labels assignment and the learned representations constantly boost each other, with more reliable pseudo-labels being discovered in each iteration.

To summarize, the contributions of this work are as follows:

\begin{itemize}
\item[1.] \textbf{Perspective:} To the best of our knowledge, our work is pioneering in the exploration of generating reliable pseudo-labels by addressing an optimal transport problem in the NID task, which is crucial for providing high-quality supervised signals.

\item[2.] \textbf{Methodology:} We propose an \texttt{EM}-optimized~\modelname{} framework for the NID problem, which iteratively enhances pseudo-labels generation and representation learning to ensure cluster-friendly intent representations.

\item[3.] \textbf{Experiments:} Extensive experiments on three challenging benchmarks show that our model establishes a new state-of-the-art performance on the NID task (average 1.5\% improvement), which confirms the effectiveness of~\modelname.

\end{itemize}
\begin{figure*}[t]
    \centering
    \includegraphics[width=0.98\linewidth]{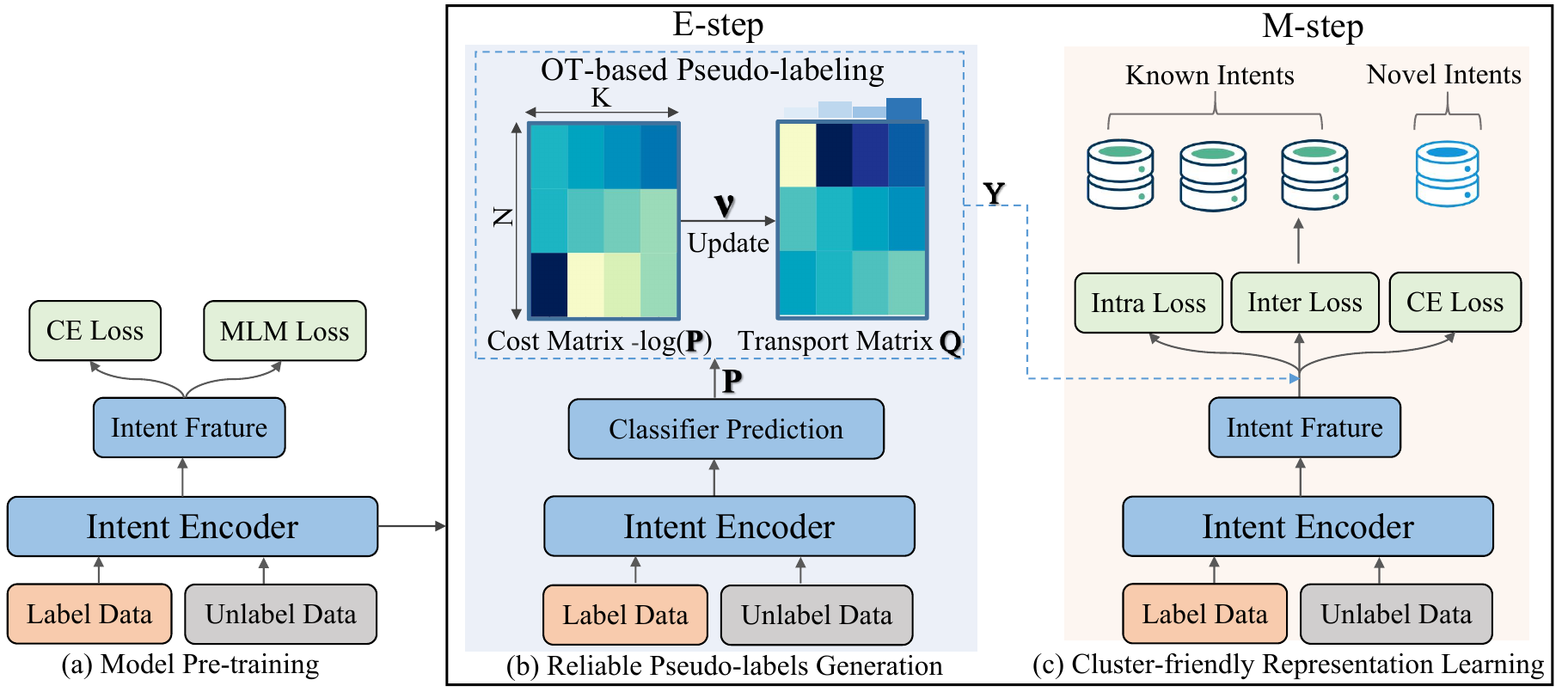}
    \caption{Overview of our~\modelname{} framework, which obtains reliable pseudo-labels by solving an optimal transport problem in \texttt{E}-step, and learns cluster-friendly representations combining Intra loss, Inter loss, and CE loss in \texttt{M}-step.}
    \label{fig:model}
\end{figure*}

\section{Our Approach}

\subsection{Problem Statement}
We introduce the setting of New Intent Discovery problem~\cite{zhang2021discovering}, which aims to recognize both known and novel intents with the aid of limited labeled intent data and unlabeled data containing all intent classes.
The dataset is composed of a labeled known classes set $\mathcal{D}_{\mathcal{L}}=\left\{x_{i}^{l}, y_{i}^{l}\right\}_{i=0}^{\left|\mathcal{D}_{\mathcal{L}}\right|}$ and an unlabeled set  $\mathcal{D}_{\mathcal{U}}=\left\{x_{j}^{u}\right\}_{j=0}^{\left|\mathcal{D}_{\mathcal{U}}\right|}$, which contains both known and novel intent classes. 
Here $x$, $y$ represents the input intent instance and the corresponding label. 
In addition, we denote the number of known and novel intent classes as $\mathcal{C}^{k}$ and $\mathcal{C}^{n}$ respectively, where $K=\mathcal{C}^{k} + \mathcal{C}^{n}$ is the total number of classes, and we assume $K$ is known~\cite{zhang2021discovering}. 
The goal is to classify known classes and cluster novel classes in $\mathcal{D}_{\mathcal{U}}$ by leveraging  $\mathcal{D}_{\mathcal{L}}$.
%
After training, model performance will be evaluated on a balanced 
testing set $\mathcal{D}_{\mathcal{T}}=\left\{x_{j}^{u}\right\}_{j=0}^{\left|\mathcal{D}_{\mathcal{T}}\right|}$.

\subsection{Approach Overview}
Figure~\ref{fig:model} illustrates the overall architecture of our proposed~\modelname{}, which employs an iterative method to bootstrap model performance through reliable pseudo-label generation and cluster-friendly representation learning.
Our method consists of three main stages. 
Specifically,
we first pre-train a feature extractor on both labeled and unlabeled data to optimize better knowledge transfer (Sec.~\ref{sec:IRTK}).
Then we obtain more accurate pseudo-labels by solving an optimal transport problem (Sec.~\ref{sec:RPGM}).
Finally, we propose intra-cluster contrastive learning and inter-cluster contrastive learning to generate well-separated clusters of representations for both known and novel intents (Sec.~\ref{sec:CRLM}).

\subsection{Intent Representation and Transferring Knowledge}
\label{sec:IRTK}
Considering the excellent generalization capability of the pre-trained model, we use the pretrained language model BERT~\cite{Devlin2019BERTPO,GanLM} as our feature extractor ($F_{\theta}:\mathcal{X} \rightarrow \mathbb{R}^{H})$. 
An overview of the intent representation and transferring knowledge (Model Pre-training) module is depicted in Fig. 2(a).
Firstly, we feed the $i^{th}$ input sentence $x_{i}$ to BERT, and take all token embeddings $[T_0, \dots, T_L]$ $\in$ $\mathds R^{(L+1) \times H}$ from the last hidden layer ($T_0$ is the embedding of the \texttt{[CLS]} token). 
The sentence representation $\boldsymbol{s}_{i} \in \mathbb R^{H}$ is first obtained by applying $\operatorname{mean-pooling}$ operation on the hidden vectors of these tokens:
\begin{MiddleEquation}
\begin{align}
\boldsymbol{s}_{i} = \operatorname{mean-pooling}([\texttt{[CLS]}, T_1,T_2,..., T_L])
\end{align} 
\end{MiddleEquation}where $\texttt{[CLS]}$ is the vector for text classification, $L$ is the sequence length, and $H$ is the hidden size. 
To effectively generalize prior knowledge through pre-training to unlabeled data, we fine-tuned BERT on labeled data ($\mathcal{D}_{\mathcal{L}}$) using the cross-entropy (CE) loss and on all available data ($\mathcal{\hat{D}} = \mathcal{D}_{\mathcal{L}} \cup \mathcal{D}_{\mathcal{U}}$) using the masked language modeling (MLM) loss. The training objective of the fine-tuning can be formulated as follows:
\begin{MiddleEquation}
\begin{equation}
     \mathcal{L}_{p} = -\mathbb{E}_{x \in \mathcal{D}_{\mathcal{L}}} \log P(y|x) - \mathbb{E}_{x \in \mathcal{\hat{D}}} \log P(\hat{x}|x_{\backslash m(x)})
\end{equation}
\end{MiddleEquation}where $\mathcal{D}_{\mathcal{L}}$ and $\mathcal{D}_{\mathcal{U}}$ are labeled and unlabeled corpus, respectively. $P(\hat{x}|x_{\backslash m(x)})$ predicts masked tokens $\hat{x}$ based on the masked sentence $x_{\backslash m(x)}$, where $m(x)$ denotes the masked tokens. The model is trained on the whole corpus $\mathcal{\hat{D}} = \mathcal{D}_{\mathcal{L}} \cup \mathcal{D}_{\mathcal{U}}$.
Through pretraining, the encoder $F_{\theta}$ can acquire both intent-specific knowledge and general knowledge from data, providing a good initialization for the representation to be further fine-tuned during subsequent training.

\subsection{Reliable Pseudo-label Generation}
\label{sec:RPGM}
To provide high-quality supervision signals for the representation learning module, we propose to generate \textit{reliable pseudo-labels} for guiding the model training, thereby transforming unsupervised training samples into pseudo-supervised samples.
Moreover, since the pseudo-labels are inevitably noisy, we design a robust learning objective to fully exploit the pseudo-supervised training samples.
An overview of the reliable pseudo-labels generation module is depicted in  Fig.~\ref{fig:model}(b), which formulates the pseudo-label assignment as an optimal transport problem.
\paragraph{Optimal Transport.}
Optimal Transport (OT) is the general problem of moving one distribution of mass to another with minimal cost.
Mathematically, given two probability vectors $\boldsymbol{\mu} \in \mathbb{R}^{m \times 1}$ and $\boldsymbol{\nu} \in \mathbb{R}^{n \times 1}$ indicating two distributions, as well as a cost matrix $\mathbf{C} \in \mathbb{R}^{m \times n}$ defined on joint space, the objective function which OT minimizes is as follows:
\begin{MiddleEquation}
\begin{equation}
\begin{gathered}
\label{eq:OT}
\min _{\mathbf{Q} \in \boldsymbol{\Pi}(\boldsymbol{\mu}, \boldsymbol{\nu})}\langle\mathbf{Q},\mathbf{C}\rangle \\ 
\boldsymbol{\Pi}(\boldsymbol{\mu}, \boldsymbol{\nu})= \left\{\mathbf{Q} \in \mathbb{R}_{+}^{m \times n} \mid \mathbf{Q} \bm{1}_{n}=\boldsymbol{\mu}, \mathbf{Q}^{\top}\bm{1}_{m}=\boldsymbol{\nu}\right\} 
\end{gathered}
\end{equation}
\end{MiddleEquation}where $\mathbf{Q}^{m \times n}$ is the transportation plan, $\langle\cdot, \cdot\rangle$ denotes frobenius dot-product, $\boldsymbol{\mu}$ and $\boldsymbol{\nu}$ are essentially marginal probability vectors. 
Intuitively speaking, these two marginal probability vectors can be interpreted as coupling budgets, which control the mapping intensity of each row and column in $\mathbf{Q}$.

\paragraph{Optimal Transport-Based Pseudo-Labels Generation (OTPL).}
In this section, we illustrate how to infer pseudo-labels \textbf{Y} via the transport matrix \textbf{Q} using the OTPL algorithm.
Given the model’s prediction $\textbf{P}\in\mathbb{R}_{+}^{N \times K}$ and its transport matrix $\textbf{Q}\in\mathbb{R}_{+}^{N \times K}$, where $N$ is the number of samples, and $K$ \footnote{We estimate the number of $K$ based on prior works~\cite{zhang2021discovering} to ensure a fair comparison.} is the number of total intent classes, OTPL considers mapping samples to classes. The cost matrix $\mathbf{C}$ can be formulated as $-\log\mathbf{P}$. 
So, we can formulate the pseudo-label generation problem as an optimal transport (OT) problem based on the problem (\ref{eq:OT}), defined as follows:
\begin{MiddleEquation}
\begin{equation}
\label{eq:OT_H}
\begin{aligned}
&\min_{\mathbf{Q}} {\langle \mathbf{Q}, {-\log{\mathbf{P}}} \rangle} + \eta \mathcal{H}(\mathbf{Q}) \\
&s.t.\,\,\mathbf{Q}\bm{1}=\boldsymbol{\mu},\mathbf{Q}^T\bm{1}=\boldsymbol{\nu}, \mathbf{Q}\geq0
\end{aligned}
\end{equation}
\end{MiddleEquation}where the function $\mathcal{H}$ is the entropy regularization, which is typically used to constrain the cluster size distribution and avoid degenerate solutions where all samples are assigned to a single cluster.
$\eta$ is a scalar factor, $\boldsymbol{\mu}=\frac{1}{N}\bm{1}$ is the sample distribution and $\boldsymbol{\nu}$ is class distribution. So the pseudo-labels matrix $\mathbf{Y}$ can be obtained by: $\mathbf{Y}=\arg \max(\mathbf{Q})$.

\paragraph{Updating Category Distribution.} We utilize model predictions to estimate class priors. 
Recognizing the potential for inaccuracies and biases during the initial stages of training, we introduce a moving-average update mechanism to bolster reliability. 
Beginning with a uniform class prior $\bm{\beta}=[1/K, \ldots, 1/K]$, we methodically refine this distribution at each epoch:
\begin{MiddleEquation}
\begin{equation}
\label{eq:Updating_Class_Distribution}
\begin{gathered}
\bm{\beta} \leftarrow \lambda_1 \cdot \bm{\beta}+(1-\lambda_1) \cdot \mathbf{b} \\
{b}_{j}=\frac{1}{N} \sum_{i=1}^{N} \mathbbm{1} \left(j=\arg \max \mathbf{P}_i \right)
\end{gathered}
\end{equation}
\end{MiddleEquation}where $\lambda_1 \in [0,1]$ is the moving-average parameter.
The class prior is consistently updated via a linear function, leading to more stable training dynamics. 
Training advancement leads to enhanced model accuracy, thus steadily improving the reliability of the estimated distribution.

\subsection{Cluster-friendly Representation Learning} 
\label{sec:CRLM}
The goal of the cluster-friendly representation learning module is to learn \textit{discriminative representations} that effectively separate known and novel intents.
Benefiting from our reliable pseudo-labels generation module, we use synthetic labels as high-quality supervisory signals to enhance discriminative representation learning.
An overview of the cluster-friendly representation learning module is shown in Fig.~\ref{fig:model}(c). We design both intra-cluster and inter-cluster contrastive learning objectives to jointly capture cluster-friendly intent representations.

\paragraph{Intra-cluster contrastive learning.} 
The Intra-cluster contrastive learning (IntraCL) method aims to acquire cluster-friendly discriminative representations by enforcing strong compactness within clusters.
We employ prototypical contrastive learning to minimize the distance between instances and their corresponding cluster prototypes, pulling instance representations closer to the matched prototypes while pushing them away from others. 
The IntraCL loss is thus defined as:
\begin{MiddleEquation}
\begin{equation}
\label{eq:intra_class}
\mathcal{L}_{a}=-\mathbb{E}_{1 \le i \le N}\log \frac{\exp \left(\boldsymbol{s}_{i} \cdot \boldsymbol\mu^{Y_{i}} / \tau\right)}{\sum_{j=1}^{K} \exp \left(\boldsymbol{s}_{i} \cdot \boldsymbol\mu^{j} / \tau\right)}
\end{equation}
\end{MiddleEquation}where the normalized sentence embeddings $\boldsymbol{s}_{i}$ matches the prototype $\boldsymbol\mu^{Y_{i}}$ of its matched label $Y_{i}$, $N$ is the number of samples and $\tau$ is a scalar temperature.
Moreover, due to the particularly volatile nature of intent representations and class cluster distributions, especially in the initial stages, acquiring a stable and ideal initial distribution is unfeasible.
Therefore, we update the intent prototypes in a moving-average style:
\begin{MiddleEquation}
\begin{gather}
\label{eq:Updating_Category_Prototypes}
{\boldsymbol\mu}_{t+1} \leftarrow \text{Normalize}(\lambda_2 \cdot {\boldsymbol\mu}_t + (1-\lambda_2) \cdot \mathbf{s}) 
\end{gather}
\end{MiddleEquation}where $\mathbf{z}$ denotes the normalized embeddings of the samples, $\boldsymbol\mu_t$ is the last updated $\boldsymbol\mu_{t+1}$ and $\lambda_2 \in [0,1]$ is the moving-average parameter.

\paragraph{Inter-cluster contrastive learning.}
We also devise a novel inter-cluster contrastive learning (InterCL) method to maximize the prototype-to-prototype distance and improve distribution uniformity by extending the instance-wise contrastive loss. 
This imposes stronger penalties on close prototypes and produces well-separated clusters. The InterCL loss is given by:
\begin{MiddleEquation}
\begin{equation}
\label{eq:inter_class}
\begin{aligned}
\mathcal{L}_{r} = 
\mathbb{E}_{1 \le i \le K} \log \frac{\sum_{j=1}^{K} \mathds{1}_{[j \neq i]} e^{s(\boldsymbol\mu_{i},\boldsymbol\mu_{j})/\tau}}{K-1}
\end{aligned}
\end{equation}
\end{MiddleEquation}where $s(\cdot,\cdot)$ is the cosine similarity to evaluate semantic similarities among prototypes.
Note that here we employ the updated $\boldsymbol\mu_{t+1}$ as modified by Eq.~\eqref{eq:Updating_Category_Prototypes}, using backpropagation to update the model, which is conducive to achieving greater separation between the prototypes.

\subsection{Optimization in~\modelname{}}
To avoid catastrophic forgetting of known categories, we also add cross-entropy loss on labeled data:
\begin{MiddleEquation}
\begin{equation}
\mathcal{L}_{\text{ce}}=-\mathbb{E}_{1 \le i \le N} y_{i}^{l} \cdot \log \left(p_{i}^{l}\right)
\end{equation}
\end{MiddleEquation}where $N$ is the mini-batch of labeled instances, $y_{i}^{l}$ is the ground-truth label of $x_{i}^{l}$. Overall, the training objective of~\modelname{} can be denoted as:
\begin{MiddleEquation}
\begin{equation}
\label{eq:final_loss}
\mathcal{L}_{all} = \alpha \mathcal{L}_{a} + (1-\alpha) \mathcal{L}_{r} + \mathcal{L}_{ce}
\end{equation}
\end{MiddleEquation}where $\alpha$ is the weight to adjust the strengths of $\mathcal{L}_{r}$ and $\mathcal{L}_{a}$.
During inference, we compute the argmax to obtain cluster assignments for testing data.

\begin{table*}[t!]
\begin{center}
\setlength\tabcolsep{6pt}
\caption{Statistics of three public datasets. \# denotes the total number of utterances. We randomly select 75\% intents as known and treat the remaining as novel intents.}
    \resizebox{0.95\textwidth}{!}{
    \begin{tabular}{l|c|c|c|c|c|c}
    \toprule
    \toprule
    Dataset & \#Classes (Known + Unknown) & \#Training & \#Validation & \#Testing & Vocabulary & Length (Max/Avg) \\
    \midrule
     CLINC & 150 (113 + 37) & 18,000 & 2,250 & 2,250 & 7,283 & 28 / 8.32  \\
     BANKING & 77 (58 + 19) & 9,003 & 1,000 & 3,080 & 5,028 & 79 / 11.91 \\
     StackOverflow & 20 (15 + 5) & 12,000 & 2,000 & 1,000 & 17,182 & 41 / 9.18 \\
    \bottomrule
    \bottomrule
    \end{tabular}
    }
    \label{tab:stastic_datasets}
\end{center}    
\end{table*}

\section{Experiments}
\subsection{Datasets}
We evaluate our proposed method on three popular intent recognition benchmarks.
The statistic of each dataset is shown in Table~\ref{tab:stastic_datasets}.
\begin{itemize}
\item[1.] \textbf{CLINC} is an out-of-scope intent classification dataset~\cite{larson2019evaluation} with 150 classes across 10 domains. As the out-of-scope utterances lack specific intent annotations for evaluation, we only utilize the 22.5K in-scope queries.
\item[2.] \textbf{BANKING} is a dataset comprising 13K customer service queries from the banking domain, spanning 77 classes. Following the data splits in~\cite{casanueva2020efficient}, we create a validation set of 1,000 utterances randomly sampled from the original training set.
\item[3.] \textbf{StackOverflow} is a dataset originally contains many technical question titles from Kaggle.com~\footnote{https://www.kaggle.com/c/predict-closed-questions-on-stackoverflow/}. In this work, we use the curated version of the dataset presented in~\cite{xu2015short}, consisting of 20K samples across 20 classes.

\end{itemize}
\subsection{Baselines}
We follow previous work~\cite{zhang2021discovering,zhou2023latent} and divide the baselines for comparison into two categories: Unsupervised (\texttt{Unsup.}) and Semi-supervised (\texttt{Semi-sup.}) methods.
\begin{itemize}
\item[1.] \textbf{Unsupervised (\texttt{Unsup.}):} 
\texttt{K-means}: KMeans with pre-trained embeddings~\cite{macqueen1967some}, 
\texttt{AC}: Agglomerative Clustering~\cite{gowda1978agglomerative}, 
\texttt{SAE-KM}: Stacked Auto Encoder and \texttt{DEC}: Deep Embedding Clustering~\cite{xie2016unsupervised}, 
\texttt{DCN}: Deep Clustering Network~\cite{yang2017towards}, 
\texttt{DAC}: Deep Aligned Clustering~\cite{Chang2017DeepAI}, 
\texttt{DeepCluster}: Deep Clustering~\cite{caron2018deep}.
\item[2.]\textbf{Semi-supervised (\texttt{Semi-sup.}):}
\texttt{KCL}: pairwise similarity predictions~\cite{hsu2017learning}, 
\texttt{CDAC+}: Constrained Adaptive Clustering~\cite{lin2020discovering}, 
\texttt{GCD}: Label Assignment with Semi-supervised KMeans~\cite{GCD}, 
\texttt{DeepAligned}: Deep Aligned Clustering~\cite{zhang2021discovering},
\texttt{DSSCC}: Deep Semi-Supervised Contrastive Clustering~\cite{DSSCC},
\texttt{DPN}: Decoupled Prototypical Network~\cite{DPN},
\texttt{CLNN}: Contrastive Learning with Nearest Neighbors and muti-task learning~\cite{zhang-2022-new-intent-discovery},
\texttt{LatentEM}: Probabilistic Framework~\cite{zhou2023latent}, 
\texttt{USNID}: Unsupervised and Semi-supervised Clustering Framework~\cite{usnid},
\texttt{DWGF}: Diffusion Weighted Graph Framework~\cite{DWGF}.
\end{itemize}
\begin{table*}[t!]
\begin{center}
\setlength\tabcolsep{6pt}
\caption{\label{tab:Main_results} Comparison against the unsupervised (\texttt{Unsup.}) and semi-supervised (\texttt{Semi-sup.}) state-of-the-art on three challenging datasets.
We bold the \textbf{best result}.}
    \resizebox{0.93\textwidth}{!}{
    \begin{tabular}{l l | c c c | c c c | c c c}
    \toprule
    \toprule
     & \multirow{3}{*}{\textbf{Methods}} & \multicolumn{3}{c}{\textbf{CLINC}} & \multicolumn{3}{c}{\textbf{BANKING}} & \multicolumn{3}{c}{\textbf{StackOverflow}} \\
    \cmidrule{3-5} \cmidrule{6-8} \cmidrule{9-11}
     &  & NMI & ARI & ACC & NMI & ARI & ACC & NMI & ARI & ACC \\
    \midrule
    \multirow{7}*{Unsup.} & K-means & 70.89 & 26.86 & 45.06 & 54.57 & 12.18 & 29.55 & 8.24 & 1.46 & 13.55 \\
    ~ & AC & 73.07 & 27.70 & 44.03 & 57.07 & 13.31 & 31.58 & 10.62 & 2.12 & 14.66\\
    ~ & SAE-KM & 73.13 & 29.95 & 46.75 & 63.79 & 22.85 & 38.92 & 32.62 & 17.07 & 34.44\\
    ~ & DEC & 74.83 & 27.46 & 46.89 & 67.78 & 27.21 & 41.29 & 10.88 & 3.76 & 13.09 \\
    ~ & DCN & 75.66 & 31.15 & 49.29 & 67.54 & 26.81 & 41.99 & 31.09 & 15.45 & 34.56\\
    ~ & DAC & 78.40 & 40.49 & 55.94 & 47.35 & 14.24 & 27.41 & 14.71 & 2.76 & 16.30 \\
    ~ & DeepCluster & 65.58 & 19.11 & 35.70 & 41.77 & 8.95 & 20.69 & 17.52 & 3.09 & 18.64 \\
      \arrayrulecolor{lightgray}
    \midrule
    \midrule
    \arrayrulecolor{black}
    \multirow{12}*{Semi-sup.} & KCL & 86.82 & 58.79 & 68.86 & 75.21 & 46.72 & 60.15 & 8.84 & 7.81 & 13.94\\
    ~ & CDAC+ & 86.65 & 54.33 & 69.89 & 72.25 & 40.97 & 53.83 & 69.84 & 52.59 & 73.48\\
    ~ & GCD & 90.54 & 66.02 & 77.15 & 72.95 & 45.70 & 58.10 & 62.37 & 45.08 & 66.90 \\
    ~ & DeepAligned & 93.95 & 80.33 & 87.29 & 79.91 & 54.34 & 66.59 & 76.47 & 62.52 & 80.26 \\
    ~ & DSSCC & 93.87 & 81.09 & 87.91 & 81.24 & 58.09 & 69.82 & 77.08 & 68.67 & 82.65 \\
    ~ & DPN & 95.11 & 86.72 & 89.06 & 82.58 & 61.21 & 72.96 & 78.39 & 68.59 & 84.23 \\
    ~ & CLNN & 96.08 & 86.97 & 91.24 & 85.77 & 67.60 & 76.82 & 81.62 & 74.74 & 86.60 \\
    ~ & LatentEM  & 95.01 & 83.00 & 88.99 & 84.02 & 62.92 & 74.03 & 77.32 & 65.70 & 80.50 \\ 
    ~ & USNID  & 96.55 & 88.43 & 92.18 & 87.53 & 69.88 & 78.75 & 80.94 & 75.08 & 86.43 \\
    ~ & DWGF  & 96.89 & 90.05 & 94.49 & 86.41 & 68.16 & 79.38 & 81.73 & 75.30 & 87.60 \\
     \arrayrulecolor{lightgray}
    \midrule
    \midrule
    \arrayrulecolor{black}
    ~ &\textbf{\modelname{}} & \textbf{97.07} & \textbf{90.93} & \textbf{95.20} & \textbf{87.16} & \textbf{68.93} & \textbf{79.48} & \textbf{84.66} & \textbf{79.51} & \textbf{89.40} \\
    \bottomrule
    \bottomrule
    \end{tabular}
    }
\end{center}
\end{table*}

\subsection{Evaluation Metrics}
We adopt three standard clustering performance metrics for evaluation~\cite{zhang2021discovering}: 
(\texttt{i}) Normalized Mutual Information \textbf{(NMI)} measures the normalized mutual dependence between the predicted labels and the ground-truth labels.
(\texttt{ii}) Adjusted Rand Index \textbf{(ARI)} measures how many samples are assigned properly to different clusters.
(\texttt{iii}) Accuracy \textbf{(ACC)} is measured by assigning dominant class labels to each cluster and taking the average precision.



\subsection{Implementation Details}
To evaluate the effectiveness of~\modelname{} on three challenging datasets, we split the datasets into train, valid, and test sets, and randomly select 10\% of training data as labeled and choose 75\% of all intents as known.
We employ the pre-trained 12-layer bert-uncased \texttt{BERT} model\footnote{\url{https://huggingface.co/bert-base-uncased}} \cite{Devlin2019BERTPO} as the backbone encoder in all experiments and only fine-tune the last transformer layer parameters to expedite the training process. We adopt the \texttt{AdamW} optimizer with a learning of 1e-5.
For all experiments, we set the temperature scale $\tau$ = 0.07 in Eq.~\eqref{eq:intra_class} and Eq.~\eqref{eq:inter_class}, the momentum factor $\lambda_1$ = 0.95 in Eq.~\eqref{eq:Updating_Class_Distribution} and $\lambda_2$ = 0.99 in Eq.~\eqref{eq:Updating_Category_Prototypes}, the scalar factor $\eta=0.05$ in Eq.~\eqref{eq:OT_H}, the weight factor $\alpha=0.7$ in Eq.~\eqref{eq:final_loss}.
For \texttt{MTP-CLNN}, the external dataset is not used as in other baselines, the parameter of top-k nearest neighbors is set to {\{100, 50, 500}\} for \texttt{CLINC}, \texttt{BANKING}, and \texttt{StackOverflow}, respectively, as utilized in \texttt{CLNN}~\cite{zhang-2022-new-intent-discovery}.
All experiments are conducted on 4 Tesla V100 GPUs and averaged over 10 runs.


\begin{table*}[t!]
\begin{center}
\setlength\tabcolsep{6pt}
\caption{\label{tab:ablation_Results} Ablation results of different model variants on three validation sets.}
    \resizebox{0.83\textwidth}{!}{
    \begin{tabular}{l l | c c c | c c c | c c c}
    \toprule
    \toprule
     & \multirow{3}{*}{\textbf{Methods}} & \multicolumn{3}{c}{\textbf{CLINC}} & \multicolumn{3}{c}{\textbf{BANKING}} & \multicolumn{3}{c}{\textbf{StackOverflow}} \\
    \cmidrule{3-5} \cmidrule{6-8} \cmidrule{9-11}
     &  & NMI & ARI & ACC & NMI & ARI & ACC & NMI & ARI & ACC \\
    \midrule
    &\textbf{\modelname{}} & \textbf{97.07} & \textbf{90.93} & \textbf{95.20} & \textbf{87.16} & \textbf{68.93} & \textbf{79.48} & \textbf{84.66} & \textbf{79.51} & \textbf{89.40} \\
     \arrayrulecolor{lightgray}
    \midrule
    \midrule
    \arrayrulecolor{black}
    & \texttt{\modelname{}-OT} & 96.94 & 89.00 & 92.84 & 86.67 & 67.82 & 76.88 & 84.20 & 76.13 & 84.20 \\
    & \texttt{\modelname{}-Intra} & 96.57 & 89.38 & 93.96 & 87.01 & 68.70 & 79.16 & 83.17 & 77.09 & 86.50 \\
    & \texttt{\modelname{}-Inter} & 96.73 & 90.05 & 94.62 & 87.22 & 69.10 & 78.67 & 84.11 & 78.43 & 87.93 \\
    \bottomrule
    \bottomrule
    \end{tabular}
    }
\end{center}
\end{table*}
\subsection{Main Results}
We present the main performance comparison results in Table~\ref{tab:Main_results}, where the best results are highlighted in \textbf{bold}. 
Overall, our proposed~\modelname{} achieves significant improvements compared with the previous strong baselines across all datasets. 
We present the result analyses from the following two aspects:

\paragraph{Comparison of different methods.}
It is observed that \modelname{} achieves the overall best performances compared to the unsupervised and semi-supervised strong baselines.
Specifically, under the average metrics across the three datasets, \modelname{} surpasses the previous state-of-the-art with an increase of 1.0\% in \texttt{ACC}, 1.3\% in \texttt{NMI}, and 2.1\% in \texttt{ARI}.
Moreover, an interesting observation is that DWGF obtains impressive performance on the BANKING dataset, but is still inferior to our proposed method in most cases.
These observations clearly validate the superiority of our method to identify known and novel intents through \textit{reliable
pseudo-label generation} and \textit{cluster-friendly representation learning}.

\paragraph{Comparison of different datasets.} We also validate the effectiveness of \modelname{} across three challenging datasets.
Specifically, 
Regarding the diverse CLINC cross-domain dataset, \modelname{} surpasses the previous strong baselines, demonstrating a 0.2\% increase in \texttt{NMI}, 0.9\% in \texttt{ARI}, and 0.7\% in \texttt{ACC}.
For the StackOverflow dataset with a smaller number of categories, \modelname{} demonstrates its effectiveness with significant improvements of 1.8\% in \texttt{ACC}, 2.9\% in \texttt{NMI}, and 4.2\% in \texttt{ARI} on average, consistently delivering substantial performance gains.
When applied to the specific single-domain BANKING dataset, \modelname{} reliably achieves significant performance improvements, underscoring its effectiveness in narrow-domain scenarios with indistinguishable intents.
These comparisons highlight the effectiveness of~\modelname{} in both coarse-grained and fine-grained datasets, making it more suitable for real-world NID scenarios.


\section{Analysis}
%
\paragraph{Ablation Study.} To assess the contribution of each component within \modelname{}, we conduct ablation experiments across three datasets in Table~\ref{tab:ablation_Results}.
Specifically,
(i) \texttt{\modelname{}-OT}, which first uses k-means to generate pseudo-labels, and then utilizes both pseudo-supervised intra-cluster and inter-cluster contrastive learning to learn representations.
The result demonstrates the effectiveness of generating accurate pseudo-labels by solving an optimal transport problem to provide high-quality supervised signals.
(ii) \texttt{\modelname{}-Intra}, which adopts only pseudo-supervised inter-cluster contrastive learning with OT-generated pseudo-labels.
The result underscores the importance of minimizing the instance-to-prototype distances to enhance within-cluster compactness.
(iii) \texttt{\modelname{}-Inter}, which implements exclusively pseudo-supervised intra-cluster contrastive learning with OT-generated pseudo-labels.
The result highlights the importance of increasing the dispersion between clusters for optimal new intent detection (NID) performance. 
Prior methods that do not explicitly constrain the distances between prototypes hinder the model's ability to learn cluster-friendly representations.
(iv) \texttt{\modelname{}}, integrating both intra-cluster and inter-cluster contrastive learning with OT-generated pseudo-labels, achieves the best performance.
These comprehensive results underline that each component of our \modelname{} plays a pivotal role in enhancing the overall effectiveness for NID.

\begin{figure*}[t]
    \centering
    \subfigure[CLINC]{
    \includegraphics[width=0.3\columnwidth]{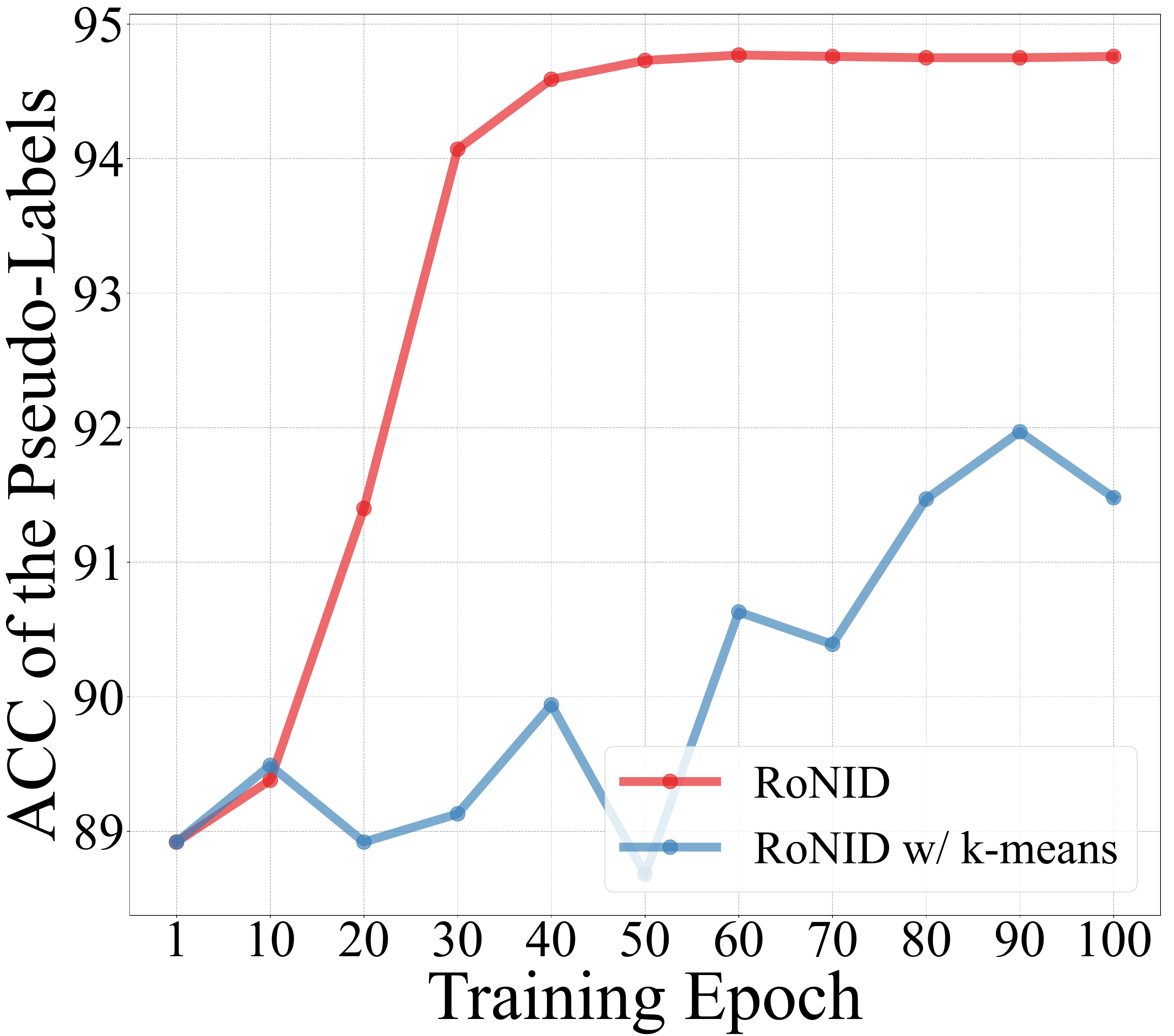}
    \label{clinc_pseudo_tsne_1}
    }
    \subfigure[BANKING]{
    \includegraphics[width=0.3\columnwidth]{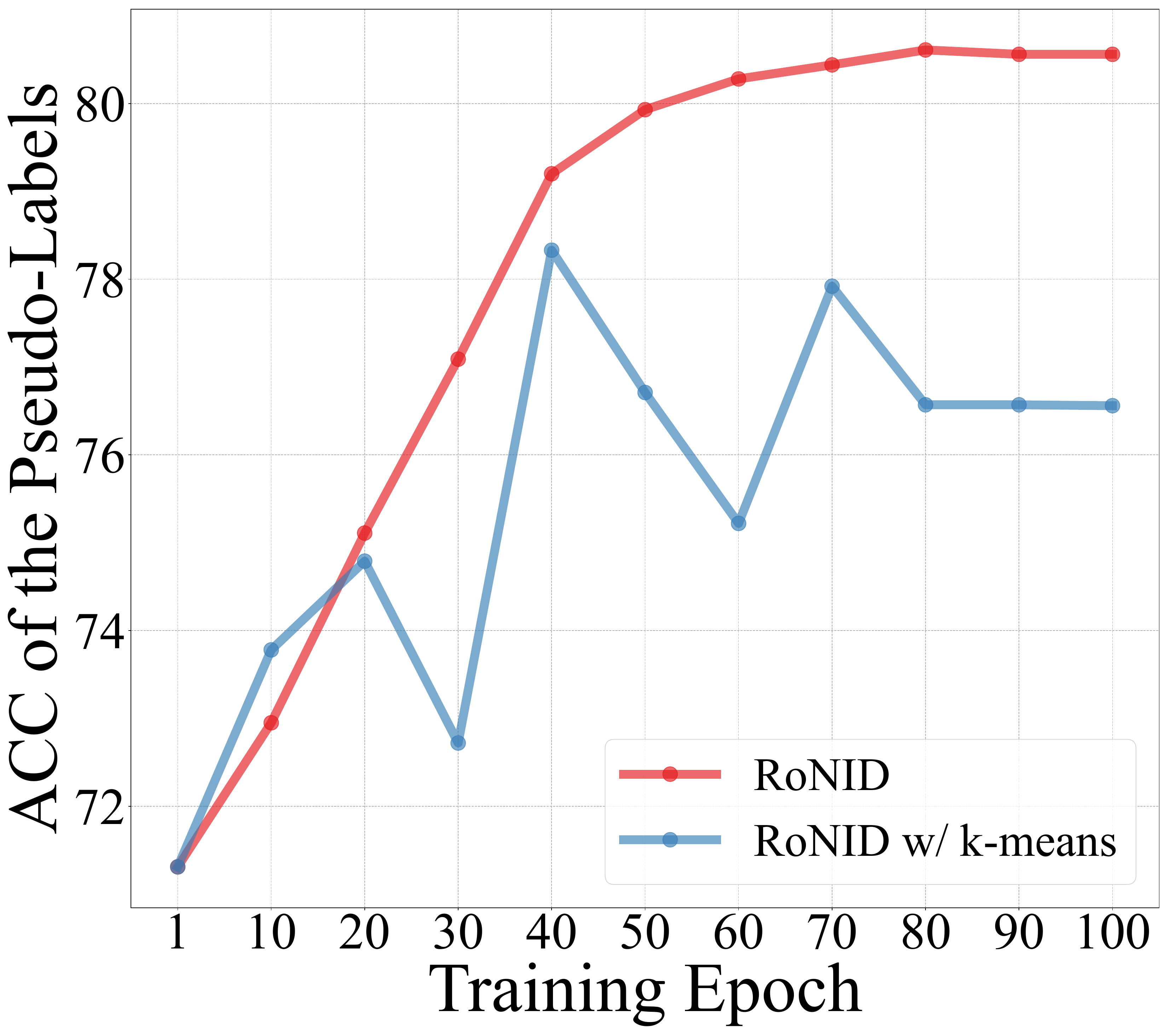}
    \label{clinc_pseudo_tsne_2}
    }
    \subfigure[StackOverflow]{
    \includegraphics[width=0.3\columnwidth]{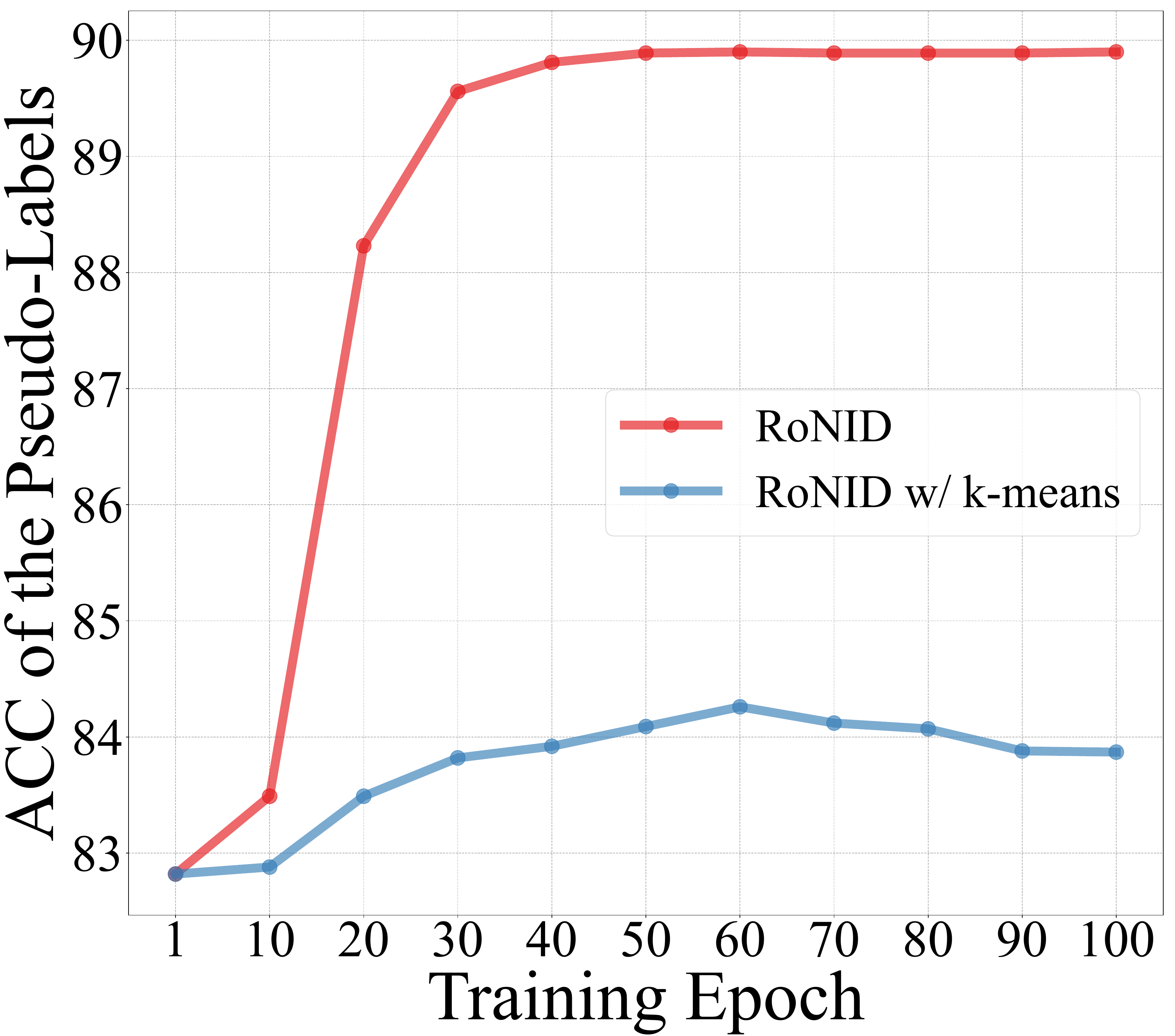}
    \label{clinc_pseudo_tsne_3}
    }
    \caption{Accuracy curves of pseudo-labels for three datasets during training. The x-axis represents training epochs, and the y-axis represents the accuracy of the pseudo-labels.} 
    \label{fig:Accuracy_curves_pseudo-label}
\end{figure*}

%

\paragraph{Effect of Pseudo-label Assignment.} 
To further investigate the contribution of the pseudo-label assignment method, we conduct a comparative analysis between our OTPL (\texttt{\modelname{}}) and the k-means (\texttt{\modelname{} w/ k-means}) strategy~\cite{zhang2021discovering,zhang-2022-new-intent-discovery,DPN,zhou2023latent}.
Throughout the training process, we meticulously evaluate the accuracy of pseudo-labels produced by both methods, aiming to precisely measure their effectiveness.
As shown in Fig.~\ref{fig:Accuracy_curves_pseudo-label}, our method demonstrates notable enhancements and achieves faster convergence compared to the \texttt{k-means} strategy across three datasets.
We hypothesize that the effectiveness of our pseudo-label assignment method stems from its well-defined and unified optimization objective. 
This is achieved by addressing an optimization transport problem, which constrains the distribution of the generated pseudo-labels to closely align with the estimated class distribution.
These findings show the efficacy of the~\modelname{} in effectively generating accurate and reliable pseudo-labels.
%
\begin{figure*}[t]
    \centering
    \subfigure[\modelname{}-Intra]{
    \includegraphics[width=0.3\columnwidth]{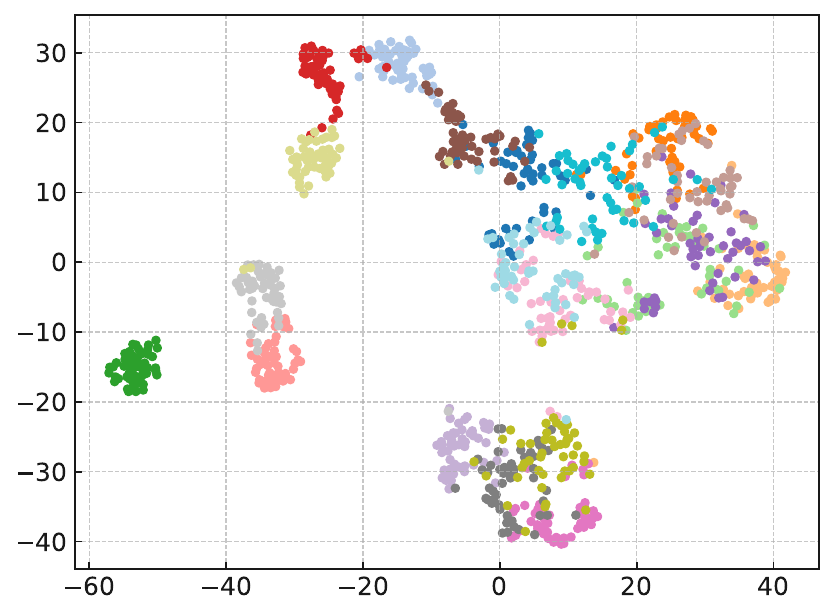}
    \label{dasfaa_tsne_1}
    }
    \subfigure[\modelname{}-Inter]{
    \includegraphics[width=0.3\columnwidth]{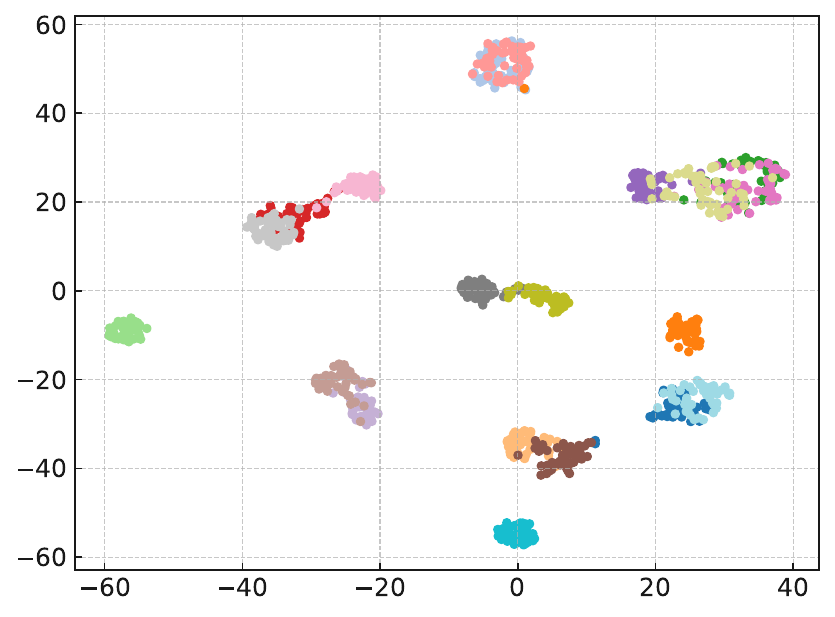}
    \label{dasfaa_tsne_2}
    }
    \subfigure[\modelname{}]{
    \includegraphics[width=0.3\columnwidth]{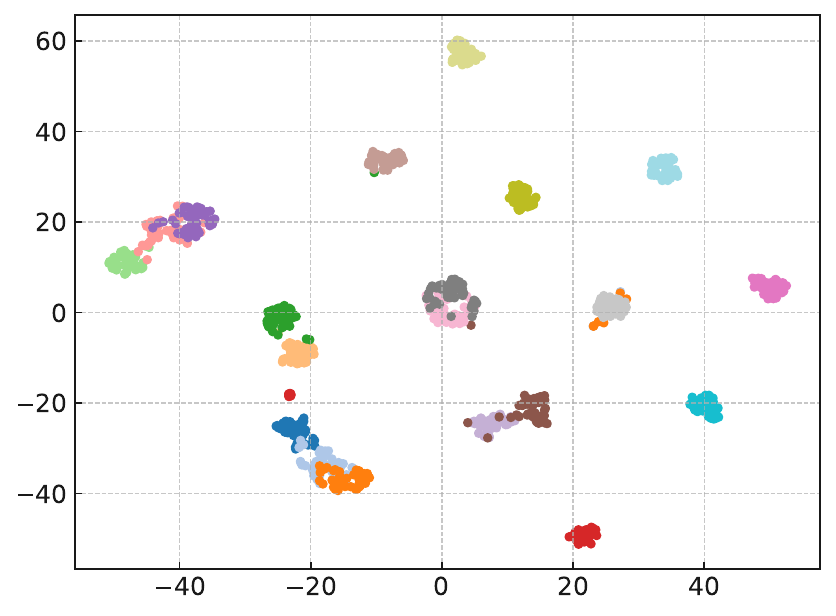}
    \label{dasfaa_tsne_3}
    }
    \caption{The t-SNE visualizations of learned intent representation.} 
    \label{fig:t-SNE}
\end{figure*}

\paragraph{Representation Visualization.} 
To assess the effectiveness of our approach in learning discriminative intent representations, we utilize the \texttt{t-SNE} to visualize projected representation on the StackOverflow dataset.
Examining Fig.\ref{dasfaa_tsne_1} and Fig.\ref{dasfaa_tsne_3}, it is evident that IntraCL effectively draws instances closer to their respective prototypes, achieving strong within-cluster compactness.
Furthermore, the contrast between Fig.\ref{dasfaa_tsne_2} and Fig.\ref{dasfaa_tsne_3} demonstrates that InterCL notably separates prototypes from each other and establishes clear cluster boundaries.
In conclusion, \modelname{} produces well-separated clusters and more distinguishable representations, which validates its effectiveness in delineating cluster boundaries for distinguishing known and novel intent classes.

\begin{figure*}[t]
    \centering
    \subfigure[Effect on CLINC]{
    \includegraphics[width=0.3\columnwidth]{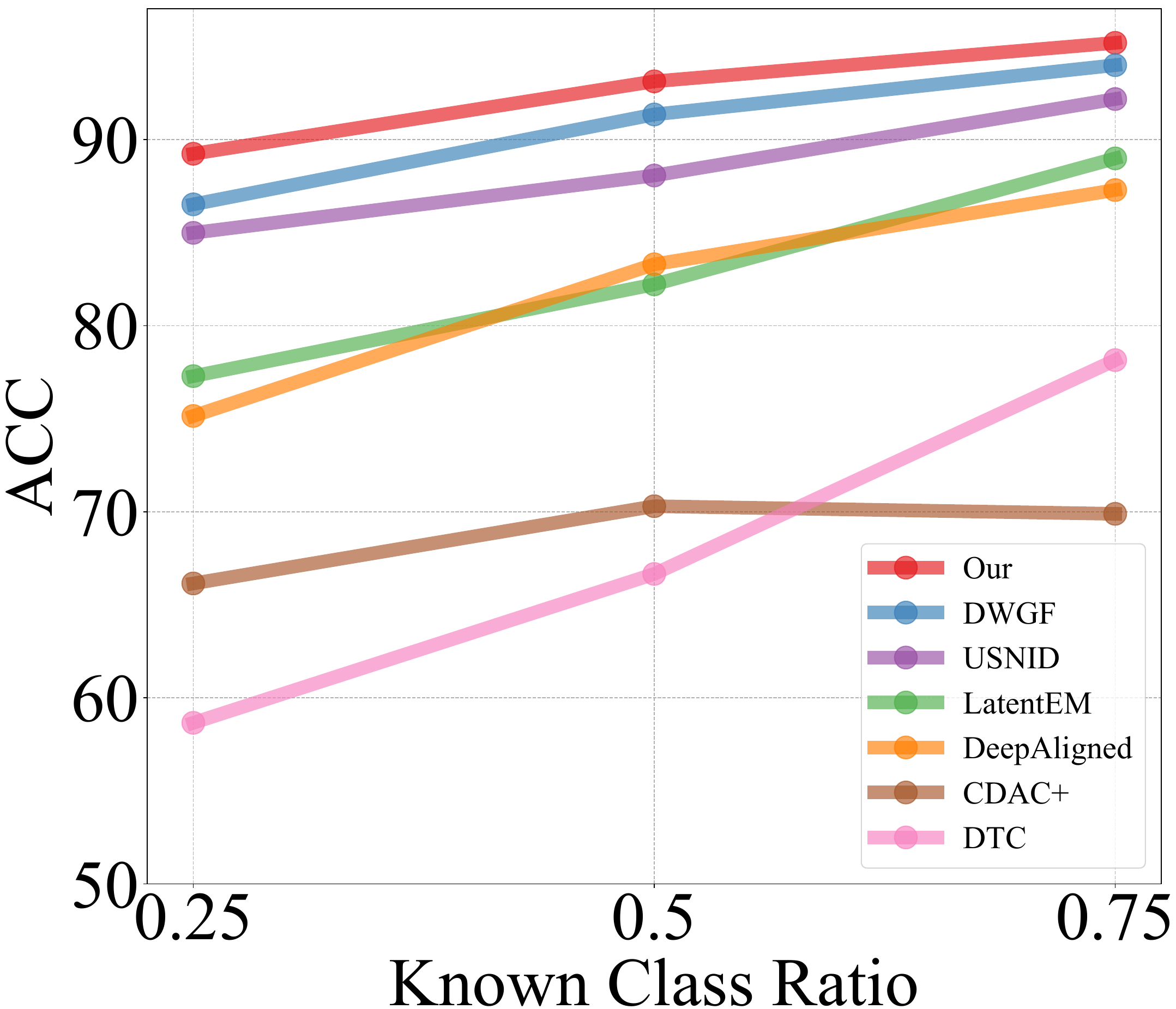}
    \label{_Clinc_ACC}
    }
    \subfigure[Effect on BANKING]{
    \includegraphics[width=0.3\columnwidth]{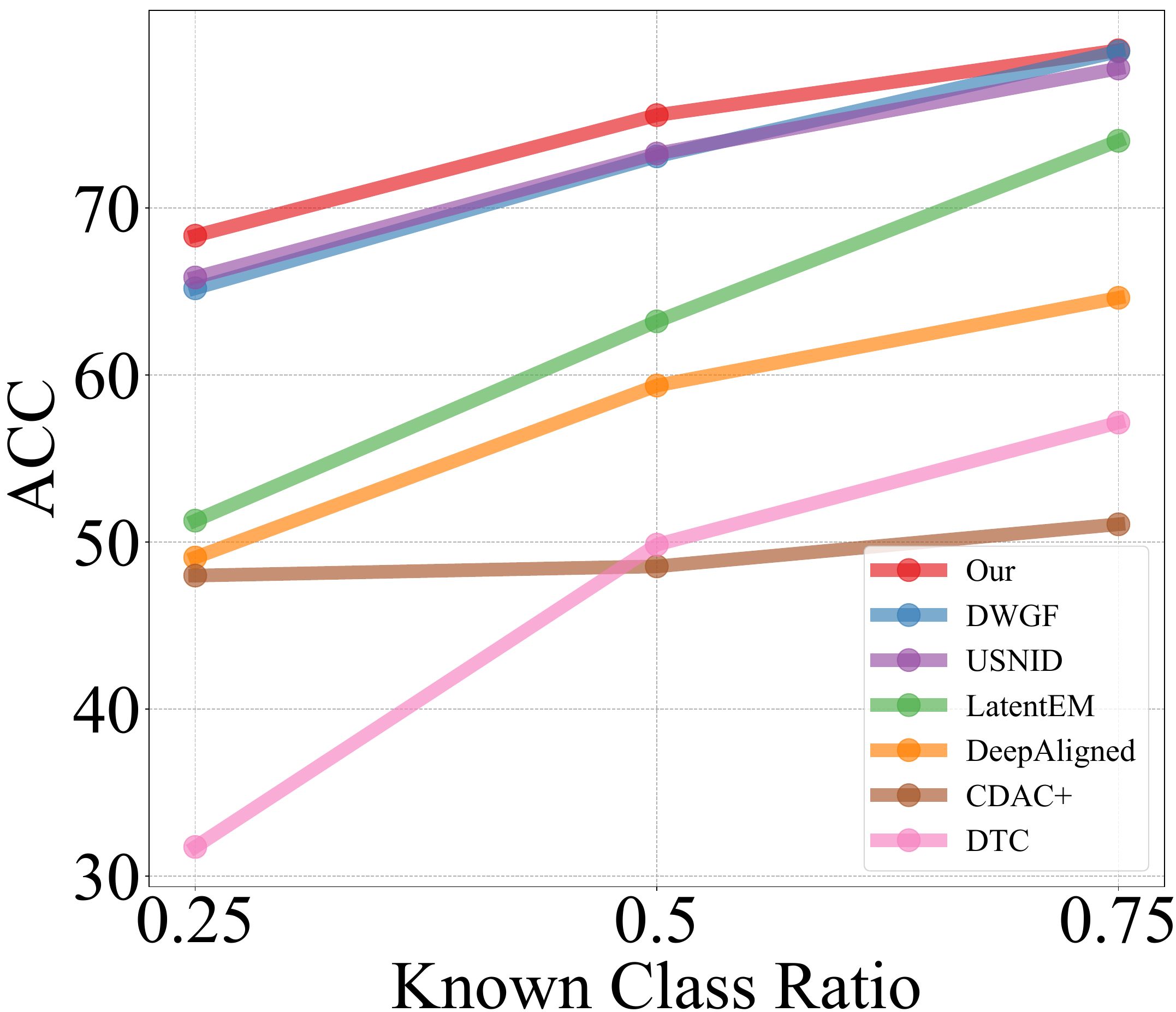}
    \label{_banking_acc}
    }
    \subfigure[Effect on StackOverflow]{
    \includegraphics[width=0.3\columnwidth]{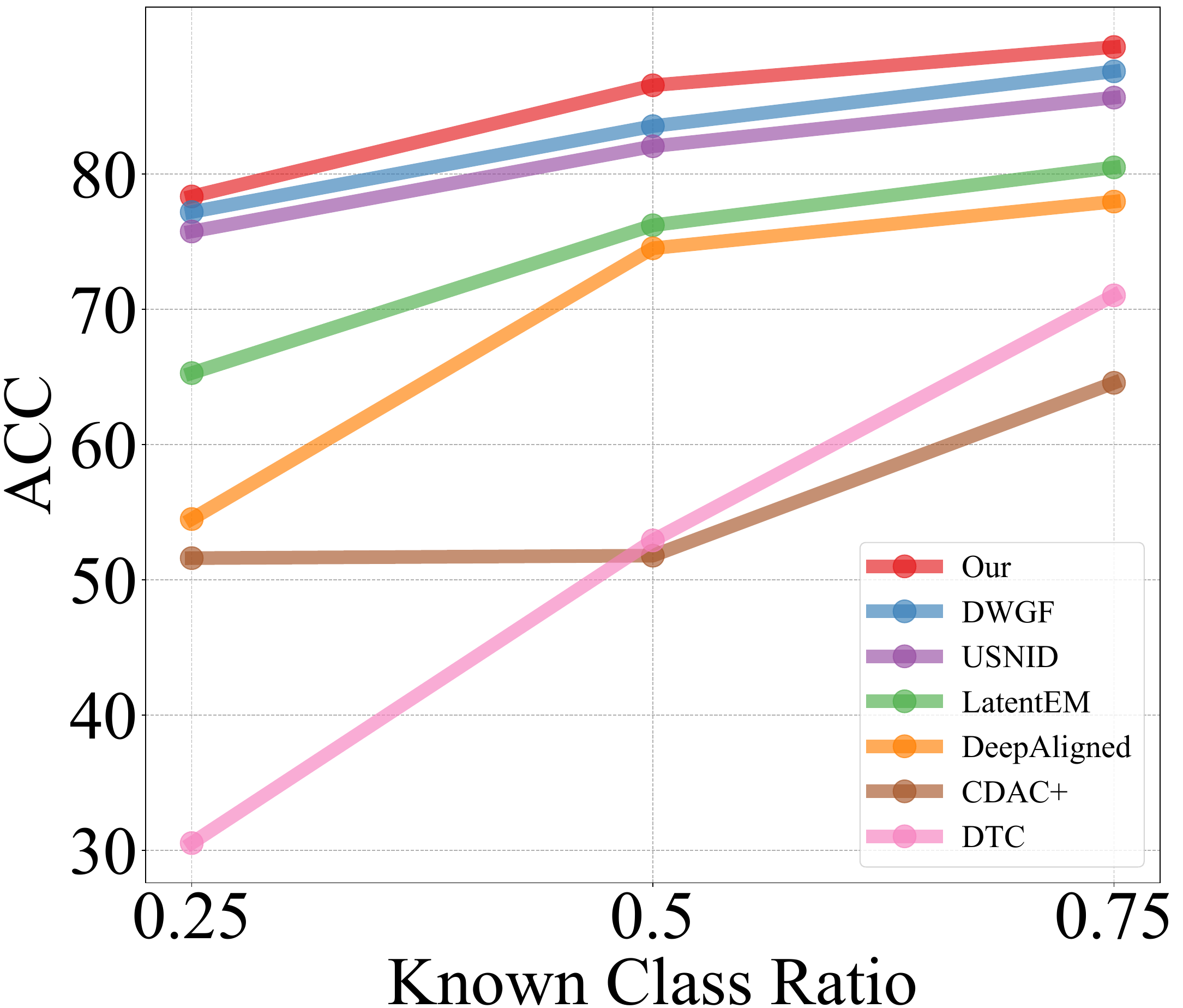}
    \label{_stack_acc}
    }
    \caption{The effect of the known class ratio on three datasets. The x-axis represents the ratio of known intent classes, and the y-axis represents the accuracy values.}
    \label{fig:known_class_ratio}
\end{figure*}

\paragraph{Effect of Known Class Ratio.}
We explore the impact of varying known intent ratios on performance by experimenting with different known class ratios (25\%, 50\% and 75\%). 
Results are presented in Fig.~\ref{fig:known_class_ratio}, our method achieves comparable or best performance under different settings on all evaluation metrics, which fully demonstrates the effectiveness and robustness of~\modelname{}.
We notice that all semi-supervised methods are affected by the number of known intents. This is likely due to less labeled data being available for guiding model training, which complicates the transfer of prior knowledge needed for discovering new intents. In contrast, as the known intent rate decreases, our proposed \modelname{} shows increasingly significant improvements.
These observations underscore that our method achieves more robust results when there are fewer known intents.

\section{Related Work}
\label{Related_Work}
\paragraph{New Intent Discovery.} New intent discovery (NID) similar to generalized category discovery (\texttt{GCD})~\cite{GCD} originating from computer vision, which aims to discover known and infers novel intents with limited labeled data and extensive unlabeled data. 
\texttt{CDAC+}~\cite{lin2020discovering} and \texttt{DeepAligned}~\cite{zhang2021discovering} learn the discriminative representation by generalizing prior knowledge to the representation of unlabeled data so that samples with similar representations can be classified into the same cluster. 
Along this line, \texttt{DSSCC}~\cite{NAACL2022}, \texttt{CLNN}~\cite{zhang-2022-new-intent-discovery}, and \texttt{USNID}~\cite{usnid} adopt different contrastive learning strategies during the pre-training phase and the clustering stage.
To further learn discriminative representations,  \texttt{LatentEM}~\cite{zhou2023latent} introduces a principled probabilistic framework, while \texttt{DPN}~\cite{DPN} proposes a decoupled prototypical network.
Recently, \texttt{DWGF}~\cite{DWGF} proposes a diffusion weighted graph framework to capture both semantic similarities and structure relationships inherent in data, enabling more sufficient and reliable supervisory signals.
\texttt{RAP}~\cite{RAP_zhang} proposes a robust and adaptive prototypical learning framework for globally distinct decision boundaries for distinguishing intent categories.
%
By contrast, we explore the generation of high-quality pseudo-labels and cluster-friendly representations to more effectively distinguish between known and novel class intents.

\paragraph{Optimal Transport.} Optimal Transport (OT) is a constrained optimization problem that aims to find the optimal coupling matrix to map one probability distribution to another while minimizing the total cost~\cite{OT_2017_Ishaan}.
Recently, Optimal Transport (OT)-based methods allow us to incorporate prior class distribution into pseudo label generation. Therefore, it has been used as a pseudo-label generation strategy for a wide range of machine learning tasks, including generative model~\cite{OT_2017_Ishaan}, semi-supervised learning~\cite{OT_2021_Kai,OT_2020_Fariborz}, clustering~\cite{OT_2020_Mathilde_Caron,OT_long_tail} and self-labelling~\cite{asano2020self}.
In this work, we explore generating reliable pseudo-labels by solving an optimal transport problem, which provides high-quality supervised signals for intent representation learning.

\paragraph{Contrastive Learning.} 
Contrastive Learning (CL) has been widely adopted to generate discriminative sentence representations for various scenarios~\cite{selfSCL,SCL2020,PCL_ICLR}, such as out-of-domain detection~\cite{DASFAA_zhang,ICASSP_zhang}, machine translation~\cite{alm,m3p,xmt,um4,wmt2021}, and named entity recognition~\cite{crop,mclner,wang2023mt4crossoie}. 
In a nutshell, the primary intuition behind CL is to pull together positive pairs in the feature space while pushing away negative pairs. 
Motivated by its superior performance, contrastive learning has also been leveraged for intent recognition in recent works~\cite{zhang-2022-new-intent-discovery,DPN,zhou2023latent,usnid,RAP_zhang}, where it is used for new intent discovery (NID).
In this work, we design both intra-cluster and inter-cluster contrastive learning objectives to capture cluster-friendly intent representations, which are more conducive to distinguishing between known and novel intent categories.


\section{Conclusion}
\label{sec:Conclusion}
In this work, we introduce an \texttt{EM}-optimized~\modelname{} framework for the NID problem, which incorporates a reliable pseudo-label generation module and a cluster-friendly representation learning module. 
The pseudo-label generation module provides high-quality supervision by accurate pseudo-label assignment through solving an optimal transport problem in~\texttt{E}-step. 
The representation learning module develops intra- and inter-cluster contrastive learning objectives in~\texttt{M}-step to obtain discriminative representations with strong within-cluster compactness and large between-cluster separation, facilitating the distinction of both known and novel intents. 
Experimental results demonstrate~\modelname{} is effective and outperforms previous state-of-the-art methods.

\subsubsection{Acknowledgements} This work was supported in part by the National Natural Science Foundation of China (Grant Nos. U1636211, U2333205, 61672081, 62302025, 62276017), a fund project: State Grid Co., Ltd. Technology R\&D Project (ProjectName: Research on Key Technologies of Data Scenario-based Security Governance and Emergency Blocking in Power Monitoring System, Proiect No.: 5108-202303439A-3-2-ZN), the 2022 CCF-NSFOCUS Kun-Peng Scientific Research Fund and the Opening Project of Shanghai Trusted Industrial Control Platform and the State Key Laboratory of Complex \& Critical Software Environment (Grant No. SKLSDE-2021ZX-18).
%
%

\bibliography{reference}
\bibliographystyle{splncs04}

\end{CJK*}
\end{document}